# SUPERPIXEL-ENHANCED PAIRWISE CONDITIONAL RANDOM FIELD FOR SEMANTIC SEGMENTATION

*Li Sulimowicz*[*]    *Ishfaq Ahmad*[*]    *Alexander Aved*[†]

[*] Department of Computer Science and Engineering, University of Texas at Arlington, TX, USA
[†] Air Force Research Laboratory, Rome, NY, USA
{*li.yin@mavs, iahmad@cse*}.*uta.edu*[*], *alexander.aved@us.af.mil*[†]

**ABSTRACT**

Superpixel-based Higher-order Conditional Random Fields (CRFs) are effective in enforcing long-range consistency in pixel-wise labeling problems, such as semantic segmentation. However, their major short coming is considerably longer time to learn higher-order potentials and extra hyperparameters and/or weights compared with pairwise models. This paper proposes a superpixel-enhanced pairwise CRF framework that consists of the conventional pairwise as well as our proposed superpixel-enhanced pairwise (SP-Pairwise) potentials. SP-Pairwise potentials incorporate the superpixel-based higher-order cues by conditioning on a segment filtered image and share the same set of parameters as the conventional pairwise potentials. Therefore, the proposed superpixel-enhanced pairwise CRF has a lower time complexity in parameter learning and at the same time it outperforms higher-order CRF in terms of inference accuracy. Moreover, the new scheme takes advantage of the pre-trained pairwise models by reusing their parameters and/or weights, which provides a significant accuracy boost on the basis of CRF-RNN [25] even without training. Experiments on MSRC-21 and PASCAL VOC 2012 dataset confirm the effectiveness of our method.

*Index Terms*— Superpixel-enhanced Pairwise CRFs, Semantic Segmentation, and Higher-order CRFs.

## 1. INTRODUCTION

Semantic segmentation is a low-level visual scene understanding problem that involves labelling the category of every pixel within images. Semantic segmentation has numerous promising applications, such as autonomous driving, robotic navigation, computer-aided medical diagnosis, and image editing [21, 11]. In recent years, the application of convolutional neural networks (CNNs) on computer vision problems has often delivered outstanding performance [19, 20, 6]. However, CNNs lack the ability to model the pixel-level correlation which can lead to "blobby" object boundaries. CRFs [13, 9, 7, 25] capture the correlation between pixels by modeling a conditional distribution between the observed and the target variables are one of the most effective and commonly used probabilistic graphical models. When CNNs are combined with CRFs [4, 25, 14, 1], they generate output with sharper and more accurate boundaries. Specifically, CRFs that are trained end-to-end with CNNs gain state-of-the-art accuracy [25, 14, 1, 3, 17].

Normally, the pairwise CRFs are not expressive enough to model higher level consistency such as region-level appearance consistency, co-occurrence of objects or detector-based cues wherein each clique consists of more than two pixels. Higher order CRFs (HO-CRFs) [7, 8, 15, 24, 1] are then used to incorporate these higher order cues, which have recently been shown to be successful in semantic segmentation [1, 18]. Conventionally, the region-level or segment-level semantic cues are formulated as higher-order potentials in two categories: (1) region-based higher-order potentials ($P^N$ Potts and robust $P^N$ Potts models [12, 7]) which are normally used as appearance consistency regular and (2) pattern-based higher-order potentials [24] which can provide independent label suggestion. In this paper, we focus on the first category and refer higher-order potentials as the region-based ones.

In the higher-order CRF, the pairwise and the higher-order term will have separate set of hyperparameters and/or parameters which need to be both learned. This results in higher cost in the parameter learning compared with pairwise models. Moreover, given the situation that there already exists a pre-trained pairwise model, updating the pairwise models to incorporate the superpixel cues, as done in H-CRF-RNN [1], requires us to fine-tune the parameters for the original pairwise term and train again for the new set of parameters of higher-order term.

To reduce the learning time of the region-based higher-order CRFs and make the above update easier, we propose an alternative approach, called superpixel-enhanced pairwise CRF, which is composed of conventional pairwise potentials and our proposed superpixel-enhanced pairwise ( SP-Pairwise) potentials. Our SP-Pairwise potentials utilized a segment filtered image as observed data to enforce the segmentation-level cues upon pairwise potentials instead. Fig. 1 illustrates the Gibbs energy structure and the segment filtered image of our superpixel-enhanced CRF. Furthermore, SP-Pairwise potential is isomorphic to the pairwise potential. This special relation leads to another important benefit, that is, the reusability of the pre-trained tunable parameters from the pairwise CRFs. Our experiments conducted on datasets MSRC-21 [23] and PASCAL VOC 2012 [5] confirm its effectiveness in terms of the accuracy and speed. The proposed method also helps reduce the amount of small spurious regions, the same as reported in [1]. Fig. 1 shows the visual comparison of the outputs.

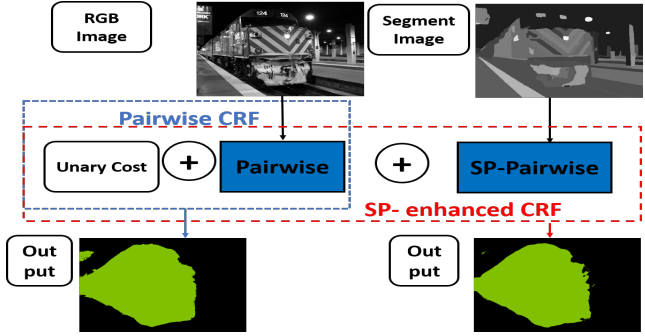

**Fig. 1**: Comparison between conventional pairwise CRFs and the proposed. The blue and red dashed box denotes the structure of the conventional CRFs and our proposed, respectively.

## 2. PRELIMINARIES

**Conditional Random Fields for Object Segmentation.** $X$ is a set of random variables, each $X_i$ corresponds to pixel at location $i$ in the image and takes value from a pre-defined set of labels $\mathcal{L} = \{l_1, ..., l_L\}$. $D$ is the observed data sequence. Let $\mathcal{G} = (\mathcal{V}, \mathcal{E})$ be a graph that $\mathcal{V} = (X_i)_{i \in \mathcal{N}}$, $\mathcal{N}$ is the total number of pixels in the given image. The conditional random field $(D, X)$ is formulated by a Gibbs distribution, where the maximum a posteriori (MAP) labelling is equivalent to $x^* = \arg\min_{x \in \mathcal{L}^N} E(X|D)$, $E(X|D) = \sum_{c \in \mathcal{C}_g} \psi_c(X_c|D)$, $\mathcal{C}_g$ is a set of cliques in graph $\mathcal{G}$.

**Pairwise CRFs.** The Gibbs energy of pairwise CRFs can be written as the sum of the unary and pairwise potentials, $E(X|D) = \sum_{i \in \mathcal{V}} \psi_i^U(x_i) + \sum_{(i,j) \in \mathcal{E}} \psi_{ij}^P(x_i, x_j)$, $(i, j) \in \mathcal{E} = i, j \in \mathcal{V}, i < j$ for the fully connected DenseCRF [9]. The unary potentials $\psi_i^U(x_i) = -\log(P(X_i = x_i))$. In this model $\mathcal{C}_g$ is the set of all unary and pairwise cliques. The pairwise potentials take the form as Eq. 1.

$$\psi_{ij}^P(x_i, x_j) = \mu(x_i, x_j) \sum_{m=1}^{K} \omega^{(m)} \kappa^{(m)}(f_i, f_j) \quad (1)$$

$\mu$ is a label compatibility function, with Potts model $\mu(x_i, x_j) = 1_{[x_i \neq x_j]}$. $\kappa^{(m)}(f_i, f_j)$ is the $m$-th Gaussian kernel. $f_i$ and $f_j$ are feature vectors for pixel $i$ and $j$. DenseCRF [9] uses contrast-sensitive two-kernel potentials, defined in terms of the color vector $I$ and position vector $P$, which is shown in Eq. 2.

$$\kappa(f_i, f_j) = \omega^{(1)} \underbrace{\exp(-\frac{|P_i - P_j|^2}{2\theta_\alpha^2} - \frac{|I_i - I_j|^2}{2\theta_\beta^2})}_{\text{appearance kernel}} \quad (2)$$

$$+ \omega^{(2)} \underbrace{\exp(-\frac{|P_i - P_j|^2}{2\theta_\gamma^2})}_{\text{smoothness kernel}}$$

here $\kappa(f_i, f_j)$ denotes $\sum_{m=1}^{K} \omega^{(m)} \kappa^{(m)}(f_i, f_j)$. The appearance kernel (bilateral kernel) forces nearby pixels with similar color to have the same labels. The degrees of the nearness and appearance similarity are controlled by parameters $\theta_\alpha$ and $\theta_\beta$. The smoothness kernel (spatial kernel) helps in removing small isolated regions.

## 3. METHOD FORMULATION

In Sec. 3.1, we present our pairwise model that incorporates superpixel cues among pairwise potentials. Then we demonstrate the details of applying this method onto fully connected CRFs in Sec. 3.2.

### 3.1. Superpixel-enhanced Pairwise CRF Model

Conventionally, the higher-order cues are formulated into higher order potentials, where each clique consists of more than two pixels. To compute such higher-order potential, it is then reformulated to a minimization problem of sum of pairwise potentials between pixel inside the higher order clique and an auxiliary random variable [12]. Moreover, the higher order term and the pairwise term each has its own set of hyperparameters and/or weights, which potentially results in longer time in learning compared with the pairwise models. Our superpixel-enhanced pairwise CRF model aims at resolving these two problems and provides an alternative other than HO-CRFs to incorporate superpixel cues.

**Formulating Superpixel Cues onto Pairwise Potentials.** Different from conventional HO-CRFs, we enforce superpixel cues on pairwise terms directly. Assume we put the pairwise graph on a segmented image, 1) we would except that pairwise potential gives out higher penalty if this pairwise edge locates inside one segment when their labels differ, we denote this potential as intra-potential, $\psi_{in}(x_i, x_j)$, and 2) a lower penalty if this edge crosses two different segments, which is denoted as $\psi_{ex}(x_i, x_j)$.

In order to meet this feature, we pre-process the original RGB images with unsupervised segmentation and use $s_i$ to denote the segment index of pixel $i$. Next, we store a segmented image wherein each pixel $i$ takes the average RGB value $C_{s_i}$ of the superpixel that it belongs to. We denote such segmented image as $D_s$ and one example is shown in Fig. 1. Suppose we take contrast-sensitive pairwise potential from [2], then our SP-Pairwise potential will be $\psi_{ij}^{SP}(x_i, x_j) = \mu(x_i, x_j)(\theta_p + \theta_v \exp(-\theta_\beta |C_{s_i} - C_{s_j}|^2))$. Here, if $s_i = s_j$ (which leads to $|C_{s_i} - C_{s_j}| = 0$), we gain maximum penalty and satisfy $\psi_{in}(x_i, x_j; s_i = s_j, \Omega) \geq \psi_{ex}(x_i, x_j; s_i \neq s_j, \Omega); \Omega = \{\theta, f_i^s, f_j^s\}$, where $f^s$ is feature from $D_s$ and $\theta = \{\theta_p, \theta_v, \theta_\beta\}$. Therefore, with $D_s$ as observed data, a contrast-sensitive pairwise potential can successfully carry out the segmentation level cues.

$$E(X|D, D_{s_1}, ..., D_{s_H}) = \sum_{i \in \mathcal{V}} \psi_i^U(x_i) + \sum_{(i,j) \in \mathcal{E}} \psi_{ij}^P(x_i, x_j; D) \quad (3)$$

$$+ \sum_{h=1}^{H} \sum_{(i,j) \in \mathcal{E}} \psi_{ij}^{SP}(x_i, x_j; D_{s_h})$$

Following this, we define the Gibbs energy of the proposed superpixel-enhanced pairwise CRF model as Eq. 3. Clearly, our Gibbs energy function has multiple terms of pairwise potentials, $\psi_{ij}^P(x_i,x_j;D)$ is the potential based on the original image and $\psi_{ij}^{SP}(x_i,x_j;D_{s_h})$ is our SP-Pairwise potential based on the segment filtered image $D_{s_h}$. Each $D_{s_h}$ can be viewed as a denoised version of $D$.

**Equivalency to Robust $P^N$ Potts Model [7].** We classify edges into intra and extra edges, $\mathcal{E} = \mathcal{E}_{in} \cup \mathcal{E}_{ex}$. We rewrite the SP-Pairwise term as $\sum_{(i,j)\in\mathcal{E}} \psi_{ij}^{SP}(x_i,x_j) = \sum_{(i,j)\in\mathcal{E}_{in}} \psi_{in}(x_i,x_j) + \sum_{(i,j)\in\mathcal{E}_{ex}} \psi_{ex}(x_i,x_j)$. Further, we decompose $\sum_{(i,j)\in\mathcal{E}_{in}} \psi_{in}(x_i,x_j)$ into the form of superpixel-based higher order potentials.

$$\sum_{(i,j)\in\mathcal{E}_{in}} \psi_{in}(x_i,x_j) = \sum_{c\in\mathcal{S}} \sum_{(i,j)\in c} \psi_{in}(x_i,x_j) \quad (4)$$

where $c$ is the superpixel clique and $\mathcal{S}$ is the set of all superpixels. Inside the superpixel clique, $|C_{s_i} - C_{s_j}| = 0$. We use $N_i(X_c)$ to denote the number of edges in the clique $c$ that have different labels, then we have Eq. 5. Let $\gamma_{max} = |c|(\theta_p + \theta_v)$. $|c|$ denotes the carnality of the pixel set $c$ which in our case is the total number of edges in the clique. We can rewrite Eq. 5 into Eq. 6.

$$\sum_{(i,j)\in c} \psi_{in}(x_i,x_j) = N_i(X_c)(\theta_p + \theta_v) \quad (5)$$

$$\sum_{(i,j)\in c} \psi_{in}(x_i,x_j) = \begin{cases} N_i(X_c)\frac{1}{|c|}\gamma_{max}, & \text{if } N_i(X_c) < |c|, \\ \gamma_{max}, & \text{otherwise.} \end{cases} \quad (6)$$

This equation proves the equivalency of the sum of SP-pairwise potentials inside each clique to a Robust superpixel-based $P^n$ model [7]. Moreover, the sum of all extra-potentials help enforce consistency between segments.

**Parameters and/or Weights Sharing.** Each pixel can be viewed as a special case of the superpixel. This isomorphism leads to the assumption that these additional superpixel-enhanced pairwise terms and the original pairwise terms can potentially share the same parameters. Thus, by formulating the $\psi_{ij}^{SP}(x_i,x_j;D_{s_h})$ with similar structure as $\psi_{ij}^P(x_i,x_j;D)$, we can avoid introducing more parameters or even weights and hence save time in learning these parameters. Moreover, if there exists a pre-trained pairwise CRF on $D$, it is possible to reuse the learned parameters or weights directly.

### 3.2. Incorporating Superpixel-enhanced Pairwise Potentials into Fully Connected CRF

For DenseCRF [22] that takes Eq. 2 as potential function, we define our fully connected SP-Pairwise potential as follows in order to reuse parameters.

$$\psi_{ij}^{SP}(x_i,x_j;D_s) = \mu_s(x_i,x_j)\omega_s^{(1)} \exp\left(-\frac{|P_i-P_j|^2}{2\theta_\alpha^{s\,2}} - \frac{|C_{s_i}-C_{s_j}|^2}{2\theta_\beta^{s\,2}}\right) \quad (7)$$

$|P_i - P_j|^2/2\theta_\alpha^{s\,2}$ acts as the weight of the color sensitive potential, thus $\theta_\alpha^s$ controls the degrees of the nearness. For the SP-Pairwise potential, we set $\theta_\beta^s = \theta_\beta, \mu_s(x_i,x_j) = \mu(x_i,x_j)$. When multiple such potential terms exist in Eq. 3, each SP-Pairwise potential term shares the same $\theta_\alpha^s, \omega_s^{(1)}$. Therefore, to incorporate different level of segment cues we only introduce one weight and one hyperparameter, which makes the whole hyperparameters to be $\{\theta_\alpha, \theta_\beta, \theta_\gamma, \theta_\alpha^s\}$ and all weights to be $\{\omega^{(1)}, \omega^{(2)}, \omega_s^{(1)}, \mu(x_i,x_j)\}$. Here $k_s(c_i^s, c_j^s)$ is used to denote this kernel.

For DenseCRF [22], we use Potts model where $\mu(x_i.x_j) = 1_{[x_i \neq x_j]}$, because the weight and the hyperparameter are all one-dimensional real value, so we do simple grid search to obtain these parameters.

For CRF-RNN [25], our SP-pairwise potential shares the same label compatibility parameter $\mu$ (a $21 \times 21$ matrix) with pairwise potential because of the isomorphism of superpixel and pixels. To update a pre-trained CRF-RNN, we set $\omega_s^{(1)} = r\omega^{(1)}, r \in (0, 1]$, $\omega^{(1)}$ is a $21 \times 21$ matrix. Thus, we only introduce two additional hyperparameters: $\theta_\alpha^s$ and $r$, which can be easily trained with grid search.

**Inference and Learning.** As done in [9, 22, 24], mean-field approximation can be used for inference in which the key step is to formulate the following iterative message passing update for different terms of potentials [10]:

$$Q_i(x_i = l) = \frac{1}{Z_i}\{-\psi_u(x_i) - \sum_{l'\in\mathcal{L}} \mu(l,l')k(f_i,f_j)Q_j(l')\} \quad (8)$$

For our SP-enhanced CRF, with simple substitution of $k(f_i, f_j)$ with $(k(f_i,f_j) + k_s(c_i^s, c_j^s))$, both the inference and learning are implemented with low complexity. The time complexity is linear with the total number of pairwise terms.

## 4. EXPERIMENTS

We evaluated our approach on two benchmarks: MSRC-21 [23] and PASCAL VOC 2012 [5] along with the baseline models, DenseCRF [9] and its recurrent neural network version CRF-RNN [25]. In our experiment, we used the accurate labelling of a subset of 92 images in **MSRC**-21 **Dataset**, which are denoted as Accurate Ground Truth (AGT). We evaluated the performance on a reduced validation set of **PASCAL VOC** 2012 **Dataset** which includes 346 images as used in [1, 25]. We used three evaluation metrics: pixel accuracy(*Global*), mean accuracy(*Average*), and mean IOU (*MeanIOU*) as used in [19] [1].

### 4.1. Evaluation on DenseCRF

We first segmented the whole set of AGT with mean-shift segmentation of two settings: $(h_s, h_r) = \{(7, 6.5), (14, 14.5)\}$, denoted as $D_{s1}$ and $D_{s2}$, respectively. Then, we split this dataset into half as training set and half as testing set. Second, we used the same unary potentials that were are in the implementation of the baseline DenseCRF [9].

---

[1] Define $n_{ij}$ as the number of pixels of class $i$ classified as $j$, $n_{cl}$ as the number of total classes in the ground truth, $t_i = \sum_j n_{ij}$ as the total number of pixels belong to class $i$, and $t$ as the number of all the pixels in an image. We have $Global = \sum_i n_{ii}/t$, $Average = \frac{1}{n_{cl}}\sum_i n_{ii}/t_i$, and $MeanIOU = \frac{1}{n_{cl}}\sum_i n_{ii}/(t_i + \sum_j n_{ji} - n_{ii})$.

We generated three models based on DenseCRF. 1) SP-CRF: we used the same energy function as DenseCRF does, except that our observation is $D_{s1}$ instead of the original image $D$. 2) DenseHO: with DenseCRF that takes $D$ as input, we add SP-Pairwise term conditioning on $D_{s1}$. 3) DenseHO2: on the basis of DenseHO, we used an additional SP-Pairwise with the coarser segments $D_{s2}$. For the higher order benchmark model, we implemented the Dense+Potts [24] as given in [24]. We use Dense+Potts to match DenseHO and Dense+Potts3 has one additional set of segments compared with DenseHO2.

Tab. 1 shows the experimental results. Our model DenseHO2 obtained 3.68% MeanIOU improvement with only 0.03$s$ additional running time compared with DenseCR, which reduced the error rate by nearly 14%. Even our simplest model SP-CRF gained around 1.7% IOU accuracy boost, which demonstrates that the segment filtered image can not only provide segment-level cues, but also partly preserve pixel-level cues. And both DenseHO and DenseHO2 have outperformed Dense+Potts on every evaluation metric using less than half of inference time consumed by Dense+Potts. This accuracy boost might be because of our model's ability to enforce appearance consistency between segments with the extra-potentials. And the faster speed is because of using pairwise potentials, the propose method benefits from the speedup of the filtered-based mean-field inference. Fig. 2 shows the examples of the visual results of these models.

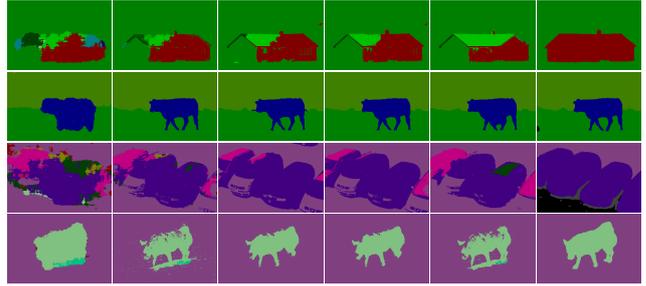

**Fig. 2**: Examples of qualitative results on MSRC-21 dataset. Columns from left to right is Unary, DenseCRF, SP-CRF, DenseHO2, Dense+Potts3, and Ground Truth.

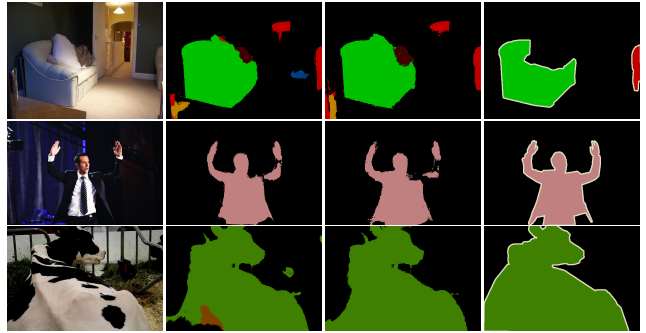

(a) Original    (b) CRF-RNN    (c) The proposed (d) Ground Truth

**Fig. 3**: Visual results on PASCAL 2010 dataset.

**Table 1**: Quantitative results evaluated on MSRC-21 dataset. For unary, it is 82.33, 83.30, 63.18 in accuracy respectively.

|  | Accurate Ground Truth | | | Time |
|---|---|---|---|---|
|  | Global | Average | IOU |  |
| DenseCRF [22] | 86.63 | 86.29 | 72.93 | 0.16$s$ |
| Dense+Potts [24] | 87.43 | 85.64 | 74.81 | 0.36$s$ |
| Dense+Potts3 [24] | 87.88 | 86.61 | 75.30 | 0.73$s$ |
| **SP-CRF** | 87.57 | 87.25 | 74.62 | 0.16$s$ |
| **DenseHO** | 87.87 | 87.82 | 76.57 | 0.19$s$ |
| **DenseHO2** | **88.20** | **88.37** | **76.61** | 0.22$s$ |

**Table 2**: Performance comparison of CRF-RNN-HO with CRF-RNN [25] and H-CRF-RNN [1].

|  | Global | Average | MeanIOU | Retrain |
|---|---|---|---|---|
| CRF-RNN [25] | **91.9** | 81.2 | 72.9 | N/A |
| H-CRF-RNN [1] | N/A | N/A | **74.0** | Yes |
| **CRF-RNN-HO** | 91.7 | **82.7** | 74.0 | No |

### 4.2. Evaluation on CRF-RNN

We segmented the reduced validation set with BGPS segmentation algorithm [16] at scale 15, denoted as $D_{s1}$, and the mean-shift segmentation with parameters set to 7, 6.5 which is denoted as $D_{s2}$. The original image set is noted as $D$. For the CRF-RNN model, we used the implementation from [25], $\theta_\alpha = 160, \theta_\beta = 3, \theta_\gamma = 3$, and $\mu, \omega^{(1)}, \omega^{(2)}$ which are $21 \times 21$ matrices that were trained end-to-end together with CNNs. Here, we constructed model CRF-RNN-HO, which includes one pairwise term and two SP-Pairwise terns with input observations as $D, D_{s1}, D_{s2}$, respectively.

We conducted simple grid search on a subset of training set in PASCAL VOC 2012 to obtain our parameters, which gives out $\theta_\alpha^s = 30, r = 0.5$. The quantitative results shown in Table 2 indicate our method gains 1.5% higher accuracy in Average and 1.1% in MeanIOU compared with CRF-RNN. Compared with H-CRF-RNN [1], we obtained equivalent performance boost. Importantly, SP-enhanced Pairwise CRF achieved the accuracy improvement without a large amount of retraining on thousands of images that is required by H-CRF-RNN. The visual results are shown in Fig. 3.

### 5. CONCLUSIONS

In the paper, we presented a novel Superpixel-enhanced pairwise CRF framework which to our knowledge is the first of such pairwise CRFs that is capable of incorporating the segment-based cues in a pixel by pixel data driven manner. We also introduced the SP-Pairwise potentials for fully connected CRF family. The results tested on semantic segmentation demonstrated that our approach improves the accuracy with easy training and is efficient in inference process. We believe this framework is generic to many other image labelling problems.